\documentclass{article} 
\usepackage[preprint]{colm2026_conference}

\usepackage{microtype}
\usepackage{hyperref}
\usepackage{url}
\usepackage{booktabs}

\usepackage{lineno}
\usepackage{graphicx}
\usepackage{booktabs}
\usepackage{amsmath}
\usepackage{amssymb}
\usepackage{tabularx}
\usepackage{multirow}
\usepackage{wrapfig}
\usepackage{tcolorbox}
\tcbuselibrary{breakable,skins}

\usepackage{xcolor}
\definecolor{qualitygreen}{HTML}{2F6F4E}   
\definecolor{knowledgeblue}{HTML}{2C4E7E}  

\newcommand{\dataset}{{\textsc{GeoAttraction-500}}\xspace}
\newcommand{\ours}{{\textsc{GAP}}\xspace}
\usepackage{xspace}

\definecolor{darkblue}{rgb}{0, 0, 0.5}
\hypersetup{colorlinks=true, citecolor=darkblue, linkcolor=darkblue, urlcolor=darkblue}

\title{Are Video Generation Models Geographically Fair? An Attraction-Centric Evaluation of Global Visual Knowledge}

\author{Xiao Liu, ~Jiawei Zhang \\
IFM Lab, Department of Computer Science\\
University of California, Davis\\
Daivs, CA, 95616, USA \\
\texttt{\{xioliu, jiwzhang\}@ucdavis.edu} \\}

\begin{document}

\ifcolmsubmission
\linenumbers
\fi

\maketitle

\begin{abstract}
Recent advances in text-to-video generation have produced visually compelling results, yet it remains unclear whether these models encode geographically equitable visual knowledge. In this work, we investigate the geo-equity and geographically grounded visual knowledge of text-to-video models through an attraction-centric evaluation. We introduce \textbf{G}eo-\textbf{A}ttraction Landmark \textbf{P}robing (\ours), a systematic framework for assessing how faithfully models synthesize tourist attractions from diverse regions, and construct \dataset, a benchmark of 500 globally distributed attractions spanning varied regions and popularity levels. \ours integrates complementary metrics that disentangle overall video quality from attraction-specific knowledge, including global structural alignment, fine-grained keypoint-based alignment, and vision--language model judgments, all validated against human evaluation. Applying \ours to the state-of-the-art text-to-video model Sora 2, we find that, contrary to common assumptions of strong geographic bias, the model exhibits a relatively uniform level of geographically grounded visual knowledge across regions, development levels, and cultural groupings, with only weak dependence on attraction popularity. These results suggest that current text-to-video models express global visual knowledge more evenly than expected, highlighting both their promise for globally deployed applications and the need for continued evaluation as such systems evolve.
\end{abstract}

\section{Introduction}

Recent advances in large-scale text-to-video models have enabled the synthesis of visually compelling and temporally coherent videos from natural language descriptions \citep{11151234, xing2024survey, ma2026controllablevideogenerationsurvey}. Models such as Kling \citep{klingai_global_2026},  Veo 3 \citep{google_aistudio_veo3_2025}, and Sora~2 \citep{openai_sora2_2025} demonstrate remarkable progress in rendering complex scenes, dynamic motion, and cinematic styles, raising expectations for their applicability across creative, educational, and informational domains. Yet, as these systems are increasingly deployed in global contexts, a critical question emerges: \emph{Do video generation models demonstrate uniform geographic and cultural knowledge, or do they privilege regions that are more prevalent in their training data?}

Prior work on AI-generated video evaluation has predominantly focused on quality aspects such as visual quality, temporal coherence, and instruction alignment, frequently utilizing automated metrics or human preference studies \citep{liu2024surveyaigeneratedvideoevaluation, he2024videoscore, zheng2025vbench20advancingvideogeneration, liu2024evalcrafterbenchmarkingevaluatinglarge, Dover, xiang2025aigvetoolaigeneratedvideoevaluation}. While these evaluations are essential, they largely overlook whether generated content accurately reflects real-world, region-specific knowledge. In parallel, studies on geo-bias and representation in Large Language Models (LLMs) \citep{Manvi2024Large, Decoupes2024Evaluation} have revealed that LLM knowledge leans toward the Global North and West \citep{lalai2025worldaccordingllmsgeographic}, exposing systematic disparities in how different regions, cultures, and languages are represented. However, analogous investigations for text-to-video models remain scarce, despite video being a richer and more demanding modality that integrates spatial structure, environmental context, and fine-grained visual details \citep{Singh2023A, chen2025t2vworldbench}. Bridging this gap is crucial not only for a deeper understanding of generative video capabilities but also for ensuring their responsible application in global contexts.

In this work, we take a step toward addressing this gap by investigating the extent of geo-equity and geographically grounded visual knowledge exhibited by text-to-video models across diverse regions. Unlike prior studies that evaluate cultural understanding through regional customs, festivals, or events \citep{chen2025t2vworldbench}, which can be stereotypical, ambiguous, and difficult to quantify objectively. We focus on representative attraction landmarks from different regions as our primary evaluation targets. Serving as globally recognizable and visually distinctive proxies for regional knowledge, these attraction landmarks are deeply tied to specific geographic identities, extensively documented in real-world imagery, and characterized by rich architectural and environmental cues. Those attraction landmarks objectively reflect the daily culture and documentary features of the region, making them uniquely well-suited for probing whether a model possesses accurate and consistent knowledge across different parts of the world \citep{Merciu2020Visual}.

To comprehensively quantify the geo-equity and geographically grounded visual knowledge, we introduce \textbf{G}eo-\textbf{A}ttraction Landmark \textbf{P}robing (\ours), the first evaluation framework that systematically assesses how faithfully a video generation model synthesizes tourist attractions from different parts of the world.

We first build a rigorous benchmark dataset, \dataset, designed to expose potential geographic disparities. Grounded in the Google Landmarks Dataset v2 \citep{ramzi2023optimization} and enriched with attraction-level metadata, our dataset comprises 500 tourist attractions distributed across diverse regions and popularity levels. This curation ensures that we can probe model performance across both well-documented global landmarks and regionally significant sites.

To assess the fidelity of generated content, we introduce a suite of complementary automatic metrics that extend beyond standard video quality evaluation. Our framework combines reference-free \citep{liu2025aigvemacs} metrics for overall visual quality with reference-based \citep{radford2021learning} metrics that specifically probe attraction-centric visual knowledge. In particular, we evaluate generated videos against real-world reference images using global structural alignment, our proposed keypoint-based local alignment metric for fine-grained geometric and textural fidelity, and VLM-as-a-Judge evaluation \citep{gu2025surveyllmasajudge}. Together, these metrics enable an objective and fine-grained assessment of how faithfully a model captures the distinctive visual identity of a geographic location.

Our comprehensive experiments not only validate the effectiveness of our proposed metrics by conducting a correlation analysis with human evaluation but also yield insightful findings through applying \ours to Sora 2 \citep{openai_sora2_2025} regarding the state of geo-equity in modern video generation. Specifically, contrary to common assumptions of severe bias toward the Global North \citep{GN} and Global West \citep{GW}, our analysis reveals that \textbf{the model maintains a relatively uniform level of visual knowledge across diverse regions and popularity levels}. Furthermore, while detailed prompts consistently improve fine-grained structural details, the model's baseline knowledge remains robust with short instructions, suggesting a stable underlying representation of global visual knowledge.

Overall, our contributions could be summarized as follows:
\begin{itemize}
    \item We propose \ours, the first systematic framework for evaluating geo-equity and globally grounded visual knowledge in text-to-video generation models, and introduce a keypoint-based metric to assess fine-grained local structural alignment.
    \item We construct \dataset, a benchmark of 500 attraction landmarks spanning diverse regions, socio-cultural contexts, and popularity levels, together with automatic metrics grounded in real-world reference image.

    \item Through experiments on Sora~2, we show that the proposed metrics align with human judgments and reveal that the model exhibits a largely uniform level of visual knowledge across regions, while remaining robust to prompt verbosity.
\end{itemize}

\section{Related Works}

\subsection{AI-Generated Video Evaluation}

AI-Generated Video Evaluation (AIGVE) is an emerging area that extends beyond traditional video quality assessment by requiring alignment with both human perception and human instructions~\citep{liu2024surveyaigeneratedvideoevaluation}. Perception-oriented methods focus on visual fidelity, realism, and temporal coherence, evolving from statistical metrics such as FVD and Inception Score~\citep{fvd,barratt2018note} to learned evaluators based on pretrained models, including DOVER for aesthetic assessment~\citep{wu_exploring_2023-1} and CLIP-based approaches such as BVQI~\citep{BVQI}. Instruction-aligned evaluation, in contrast, measures text–video consistency using metrics like CLIPScore~\citep{hessel2022clipscorereferencefreeevaluationmetric} and TIFA~\citep{hu2023tifa}, but often overlooks perceptual degradation and physical implausibility. These two lines have largely developed independently and typically reduce complex judgments to single scalar scores.

Recent work has sought to unify these aspects through multi-dimensional and model-based evaluation frameworks. VBench~\citep{huang_vbench_2023} and EvalCrafter~\citep{liu_evalcrafter_2023} aggregate multiple quality and alignment dimensions, while VideoScore \citep{he2024videoscore} and VideoScore2 \citep{he2025videoscore2thinkscoregenerative} leverage large vision–language models for holistic assessment, with the latter introducing interpretable, multi-dimensional reasoning over visual quality, text alignment, and physical consistency. AIGVE-MACS \citep{liu2025aigvemacs} further advances this direction by jointly producing aspect-wise scores and natural language comments trained on large-scale human annotations, improving interpretability and human alignment. Despite this progress, existing AIGVE methods primarily focus on overall quality and prompt adherence, and remain largely agnostic to whether generated videos faithfully reflect structured real-world or region-specific visual knowledge.

\subsection{Geo-Bias and Global Representation in Generative Models}

Geographic bias and uneven global representation have been extensively documented in LLMs. Prior studies show that LLMs systematically favor entities, knowledge, and perspectives from the Global North and Global West, exhibiting higher accuracy and richer representations for Western regions while underperforming on or misrepresenting regions from the Global South~\citep{manvi2024geobias, lalai2025world, tao2024cultural}. These disparities have been observed across tasks such as entity deduction, factual reasoning, and cultural judgment, indicating that geo-bias is a persistent and structural issue rooted in training data distributions~\citep{manvi2024geobias}.

In contrast, geo-bias in generative video models has received far less attention. Existing work such as T2VWorldBench~\citep{chen2025t2vworldbench} represents an early attempt to evaluate world knowledge in text-to-video generation, but relies largely on culturally stereotypical prompts and coarse qualitative judgments, making systematic and objective evaluation difficult. As a result, there remains no systematic, large-scale evaluation of how fairly text-to-video models represent diverse regions of the world, motivating the need for geographically grounded and structurally measurable evaluation protocols.

\section{Geo-Attraction Landmark Probing}

\begin{figure}[t!]
    \centering
    \includegraphics[width=0.95\linewidth]{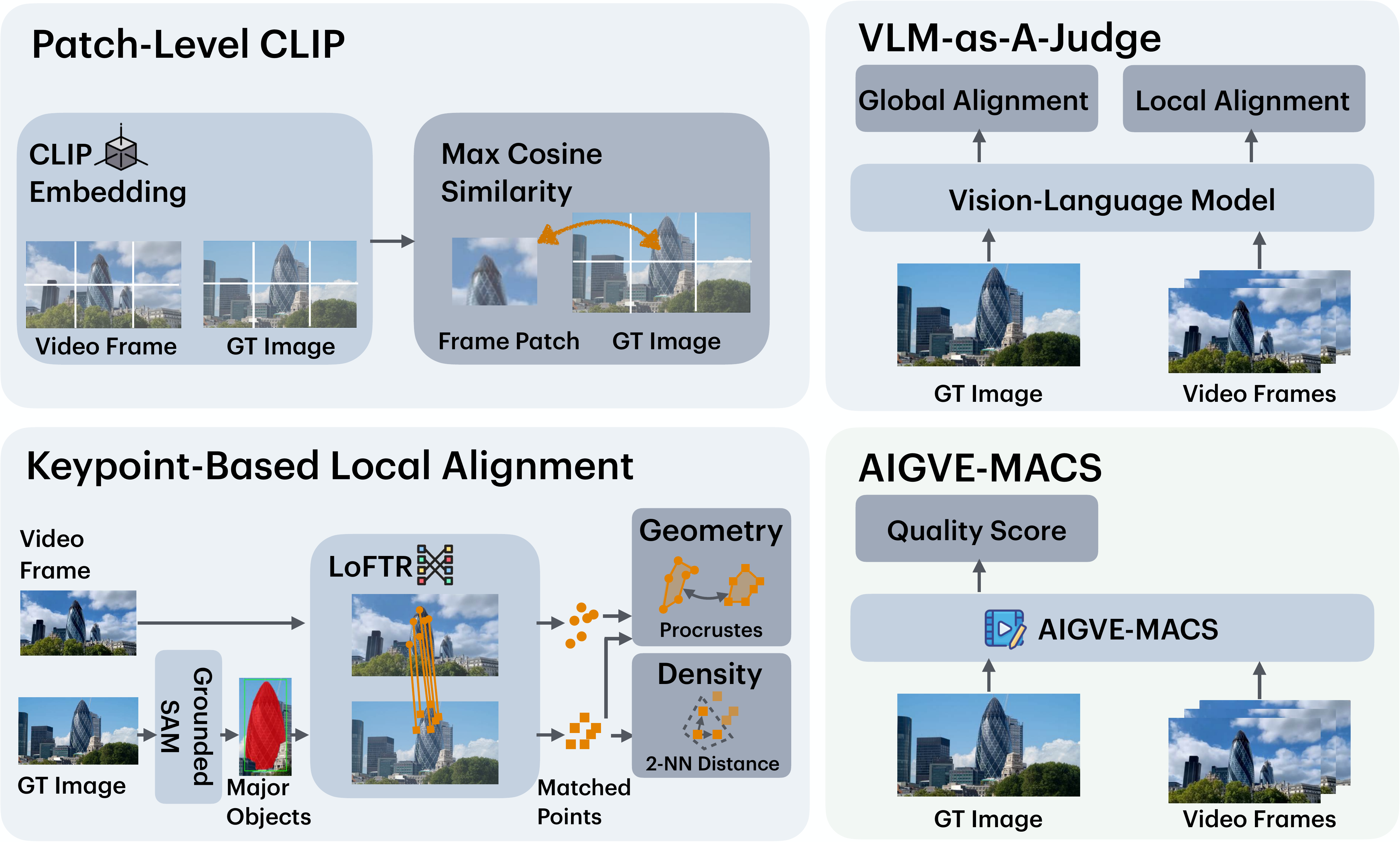}
    \caption{
Overview of the evaluation metrics used in Geo-Attraction Landmark Probing (\ours).
GAP integrates complementary \textcolor{knowledgeblue}{knowledge-oriented} and \textcolor{qualitygreen}{quality-oriented} metrics to assess generated videos.
\textcolor{knowledgeblue}{Patch-Level CLIP} measures global structural alignment via patch-wise similarity,
\textcolor{knowledgeblue}{Keypoint-Based Local Alignment} evaluates fine-grained geometric and textural consistency,
and \textcolor{knowledgeblue}{VLM-as-A-Judge} provides semantic assessments of global and local alignment.
In contrast, \textcolor{qualitygreen}{AIGVE-MACS} evaluates overall video quality independently of attraction-specific knowledge.
}
    \label{fig:main_figure}
\end{figure}

To systematically investigate the geo-equity and geographically grounded visual knowledge of text-to-video models through their visual outputs, we propose a novel evaluation framework termed \textbf{G}eo-\textbf{A}ttraction Landmark \textbf{P}robing (\ours). This framework examines whether a video generation model can faithfully synthesize tourist attractions that capture the architectural, environmental, and socio-cultural characteristics of their corresponding locations.

We focus on tourist attractions because they provide globally recognizable and visually distinctive proxies for regional knowledge. Such landmarks are closely tied to specific geographic identities, extensively documented in real-world imagery, and characterized by rich visual signals spanning architecture, landscape, and socio-cultural context. Consequently, an attraction-centric evaluation offers a controlled yet expressive setting for assessing whether a model exhibits uneven geographic knowledge.

In \ours, we prompt a video generation model to synthesize short videos of tourist attractions from specified countries or regions. Generated videos are first evaluated using AIGVE-MACS \citep{liu2025aigvemacs} to assess overall visual quality, and are then compared against real-world reference images using three complementary knowledge-oriented metrics: Patch-Level CLIP \citep{radford2021learning} for global structural alignment, our novel Keypoint-Based Local Alignment for fine-grained structural fidelity, and VLM-as-A-Judge \citep{gu2025surveyllmasajudge} for semantic alignment, as illustrated in Figure~\ref{fig:main_figure}. 

By applying a unified evaluation pipeline across geographically diverse attractions, \ours enables systematic analysis of whether text-to-video models exhibit region-dependent disparities or instead demonstrate a consistent level of globally grounded visual knowledge. In the remainder of this section, we describe the construction of \dataset benchmark dataset in Section~\ref{sec:data_cons} and detail the evaluation metrics used in \ours in Section~\ref{sec:metric}.

\subsection{\dataset Benchmark Dataset}
\label{sec:data_cons}

We construct our benchmark dataset, \dataset, based on the Google Landmarks Dataset v2 (GLDv2)\footnote{\url{https://github.com/cvdfoundation/google-landmark}} \citep{ramzi2023optimization} and augment it with attraction-level metadata from the Google Landmarks Places dataset\footnote{\url{https://huggingface.co/datasets/visheratin/google_landmarks_places}}. This combined dataset provides landmark images paired with attraction names and corresponding city, country, and regional geographical information.

To capture variation in global visibility, we incorporate Wikipedia page view statistics for each attraction landmark as a proxy for attraction popularity. Page view counts reflect how frequently an attraction landmark is accessed by users worldwide and enable analysis of model performance across attractions with varying levels of global prominence.

To ensure that each reference ground-truth image faithfully captures the defining characteristics of an attraction and its surrounding environment, we employ human annotators to manually select a representative image for each attraction from the GLDv2 dataset. These ground-truth images are chosen to reflect the canonical visual appearance of the landmark and serve as reference visuals for subsequent comparisons.

Based on each selected ground-truth image, we further generate textual instructions for video generation. Specifically, we prompt GPT-5.1 \citep{openai_gpt5.1_2025} to produce a structured textual description of the scene conditioned on the attraction’s name and location. The output includes (i) a detailed cinematic caption of 3--6 sentences describing composition, camera viewpoint, environmental and architectural details, lighting conditions, and stylistic cues, and (ii) a concise single-sentence caption summarizing the same scene. Both captions are designed to be visual, concrete, and directly suitable for scene recreation by video generation models. We present the instruction generation prompt in Appendix~\ref{sup:caption_prompt}.

\begin{figure}[t!]
    \centering
    \includegraphics[width=0.85\linewidth]{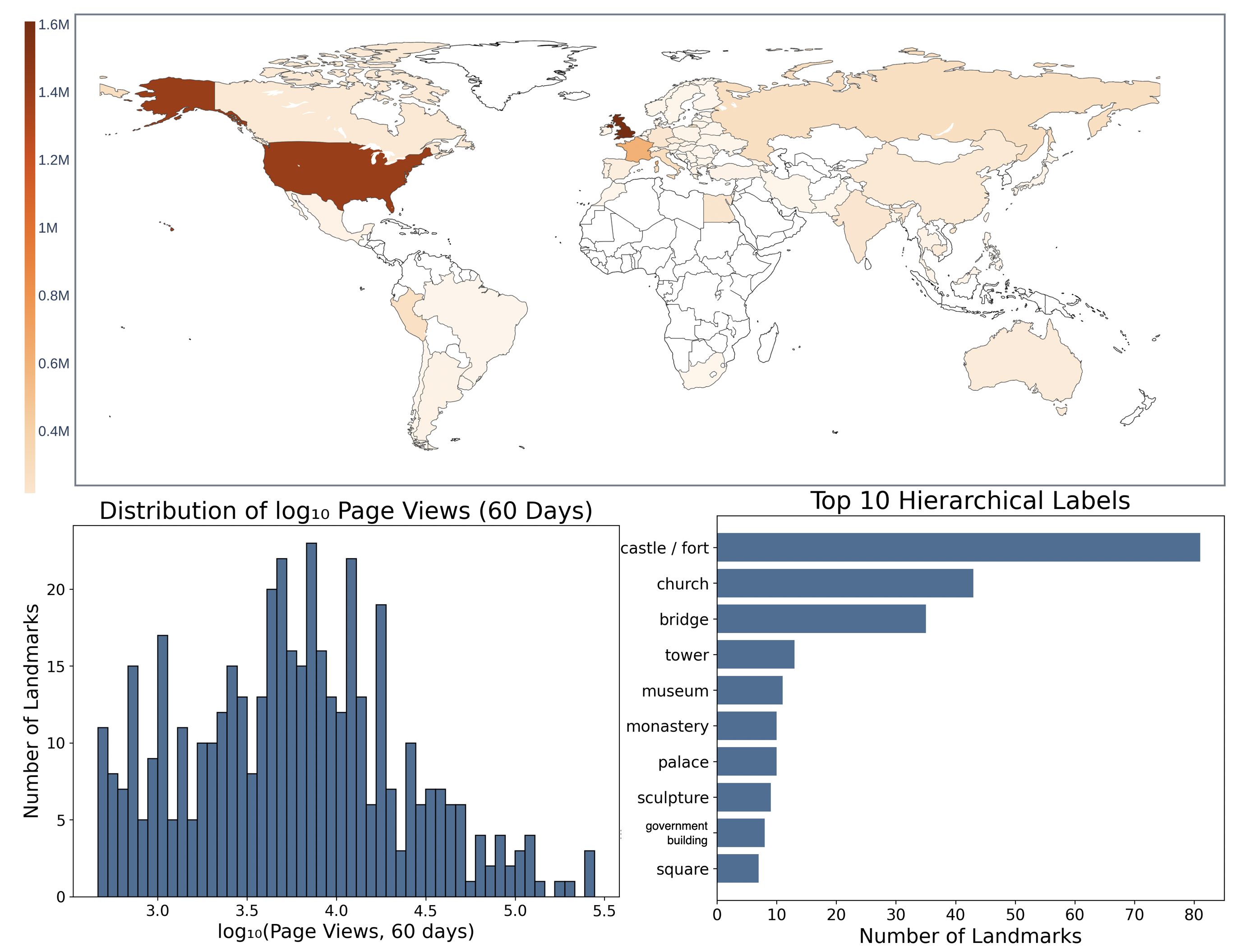}
    \caption{Dataset statistics. \textbf{Top:} Geographic distribution of attractions by country, showing broad global coverage. \textbf{Bottom left:} Distribution of pageviews of attraction landmarks in log scale. \textbf{Bottom right:} Top 10 hierarchical attraction landmark categories, dominated by architectural and cultural structures.}

    \label{fig:data_dist}
\end{figure}

In total, \dataset comprises 500 carefully curated tourist attraction landmarks distributed worldwide, covering diverse geographic regions, socio-cultural contexts, and popularity levels. Detailed statistics and the geographic distribution of attractions are presented in Figure~\ref{fig:data_dist}.

\subsection{Evaluation Metrics}
\label{sec:metric}

To comprehensively assess both the visual quality and the geographically grounded visual knowledge expressed by a video generation model, we employ a suite of complementary evaluation metrics. These metrics are designed to disentangle general perceptual quality from attraction-specific knowledge fidelity, enabling a fine-grained analysis of geo-equity across regions.

For each generated video, we uniformly sample $N{=}5$ frames $\mathcal{F}=\{f_1,\dots,f_N\}$ and compute all metrics on this shared set. Each metric aggregates frame-level scores in a manner consistent with its evaluation objective, as detailed below. The overview of each metric is presented in Figure~\ref{fig:main_figure}.

\paragraph{AIGVE-MACS}\citep{liu2025aigvemacs}.
We assess overall video generation quality using AIGVE-MACS, a multi-aspect evaluation model that scores generated videos across nine perceptual dimensions, including visual quality, coherence, and aesthetics, on an integer scale from 0 to 5. As a reference-free, quality-forwarded metric, AIGVE-MACS captures general visual fidelity independent of geographic or attraction-specific knowledge.
The score is computed by inputting all sampled frames, and we report the overall score as the video quality measure.

\paragraph{Patch-level CLIP}\citep{radford2021learningtransferablevisualmodels}.
To evaluate attraction-specific knowledge at a global structural level, we compute Patch-level CLIP similarity between sampled video frames and the ground-truth reference image of the target attraction. This metric assesses whether the generated video captures the overall spatial layout, silhouette, and environmental context of the attraction, while remaining robust to variations in viewpoint, scale, and appearance.

Given a ground-truth image and a sampled frame $f$, we partition both images into fixed-size patches following the DINO \citep{oquab2024dinov2learningrobustvisual} preprocessing pipeline and extract patch-level visual tokens. Each image is thus represented as a set of patch embeddings corresponding to local spatial regions.

Let $\mathbf{T}^{\mathrm{gt}}=\{\mathbf{t}^{\mathrm{gt}}_i\}_{i=1}^{n}$ denote the patch embeddings extracted from the ground-truth image and $\mathbf{T}^{f}=\{\mathbf{t}^{f}_j\}_{j=1}^{m}$ denote those extracted from frame $f$, where $n$ and $m$ are the numbers of patches in the ground-truth image and the video frame, respectively. All embeddings are $\ell_2$-normalized, and patch similarity is measured using cosine similarity. To account for asymmetric visual coverage between the two images, we compute a symmetric max-matching score:
\begin{equation}
\begin{aligned}
\text{Patch-CLIP}(f) = \frac{1}{2} \Bigg(
&\frac{1}{n}\sum_{i=1}^{n} \max_{j}\operatorname{cos-sim}\!\left( \mathbf{t}^{\mathrm{gt}}_i, \mathbf{t}^{f}_j \right) 
+ \frac{1}{m}\sum_{j=1}^{m} \max_{i}\operatorname{cos-sim}\!\left( \mathbf{t}^{f}_j, \mathbf{t}^{\mathrm{gt}}_i \right)
\Bigg).
\end{aligned}
\end{equation}

The Patch-level CLIP score is averaged over the sampled frames to obtain the video-level score. 


\paragraph{Keypoint-based Local Alignment.}
While Patch-level CLIP captures global structural correspondence, it does not explicitly assess fine-grained local geometric and textured fidelity. In particular, it may overlook whether localized surface details, such as architectural textures, facade patterns, and fine-scale structural elements, are faithfully reproduced. To evaluate whether a model preserves such detailed structural and textured characteristics of an attraction, we introduce a keypoint matching-based alignment metric that measures local detailed correspondence between generated frames and the ground-truth image. 

By explicitly comparing dense keypoint correspondences within landmark-relevant regions and jointly accounting for match density and geometric consistency, this metric provides a fine-grained and interpretable measure of structural alignment, which is essential for detecting subtle errors in geographically grounded visual knowledge that may be overlooked by global similarity measures.

Specifically, given the ground-truth image, we first identify $K$ landmark-relevant regions as binary masks $\{M_k\}_{k=1}^{K}$ using Grounded SAM \citep{ren2024grounded}. These regions correspond to visually salient components of the attraction, such as buildings, towers, or distinctive architectural elements. For each sampled video frame $f$, we extract dense keypoint correspondences using LoFTR \citep{sun2021loftrdetectorfreelocalfeature},
$\mathcal{P}^f=\{(p^{\mathrm{gt}}_i,p^{f}_i)\}_{i=1}^{N_f}$,
where $p^{\mathrm{gt}}_i$ and $p^{f}_i$ denote matched keypoints in the ground-truth image and the generated frame, respectively. For each region $M_k$, we define the index set
$\mathcal{I}^f_k$,
containing matches whose ground-truth keypoints fall inside the region.



To account for differences in intrinsic visual complexity across regions, we introduce a region-specific detailness term to normalize the keypoint matching process. Human-made structures often exhibit dense, fine-grained details and yield many keypoint correspondences, whereas natural attractions typically have lower structural complexity. Without normalization, regions with greater intrinsic detail could disproportionately influence the alignment score.

We compute detailness from the Laplacian response of the ground-truth image. Let $L$ denote the grayscale Laplacian, and define the detailness of region $M_k$ as
\begin{equation}
d_k = 1 - \exp\left(-\mathrm{Var}(L \mid M_k)/\tau\right),
\end{equation}
where $\tau$ is a fixed scaling constant. This normalization ensures that alignment scores reflect relative matching quality conditioned on regional visual complexity, rather than absolute keypoint density.

With the region-specific detailness defined above, we next estimate the density of matched keypoints within each region $M_k$ for frame $f$. Match density serves as a measure of how consistently and densely local structures in the ground-truth image are recovered in the generated frame. Intuitively, a higher density of well-distributed matches suggests that local surface patterns and geometric details, such as architectural textures or repeated facade elements, are faithfully reproduced, rather than matched sparsely or incidentally.

Specifically, let $\{p^{\mathrm{gt}}_i\}_{i\in\mathcal{I}^f_k}$ be the ground-truth keypoints inside the region. We compute the inverse of the mean distance to the second nearest neighbor:

\begin{equation}
\resizebox{0.32\linewidth}{!}{$
\rho^f_k =
\left(
\frac{1}{|\mathcal{I}^f_k|}
\sum_{i\in\mathcal{I}^f_k}
\mathrm{NN}_2(p^{\mathrm{gt}}_i)
\right)^{-1}
$}
\end{equation}

where $\mathrm{NN}_2(\cdot)$ denotes the Euclidean distance to the second nearest neighbor. 

To normalize for region-specific scale and intrinsic visual complexity, we divide $\rho^f_k$ by a self-matching baseline $\rho^{\mathrm{ref}}_k$, which is computed from LoFTR correspondences between the ground-truth image and itself. This self-matching normalization mitigates biases introduced by the LoFTR matcher and isolates alignment differences attributable to the generated content. The resulting density is further normalized by the region detailness to remove the influence of intrinsic regional complexity:

\begin{equation}
r_k^{f} = \frac{\rho_k^{f}}{d_k\,\rho_k^{\mathrm{ref}}}.
\end{equation}

The resulting region-level match density score is obtained via a saturating transform:
\begin{equation}
D^f_k = 1 - \exp\!\left(-r^f_k /\beta\right),
\end{equation}
where $\beta$ is a fixed constant.

In addition to match density, we assess geometric consistency by applying Procrustes analysis \citep{Andreella2020ProcrustesAF} to the matched keypoints and computing the Procrustes disparity $\delta^f_k$, which is converted into a similarity score:
\begin{equation}
G^f_k = 1 - \delta^f_k.
\end{equation}

We aggregate region-level scores using area-weighted averaging. Let $a_k = \sum M_k$ denote the area of region $M_k$, and define normalized weights
$w_k = a_k / \sum_{\ell=1}^{K} a_\ell$.
Frame-level density and geometry scores are then computed as
\begin{equation}
D(f) = \sum_{k=1}^{K} w_k D^f_k,
\qquad
G(f) = \sum_{k=1}^{K} w_k G^f_k.
\end{equation}
The overall alignment score for frame $f$ is defined as
\begin{equation}
\operatorname{Keypoint-Match}(f) = D(f) + G(f).
\end{equation}

Finally, the video-level keypoint matching-based alignment score is defined as the maximum score over the sampled frames. This max aggregation accounts for temporal variability in viewpoint and composition, reflecting the model’s ability to generate at least one frame that faithfully captures the fine-grained structure of the target attraction.

\paragraph{VLM-as-A-Judge}\citep{gu2025surveyllmasajudge}.
To complement the vision-based metrics with a semantic perspective, we employ a VLM as an automated judge. Specifically, we use GPT-5.1 to compare the ground-truth image with the sampled frames $\mathcal{F}$, conditioned on the generation instruction. The model outputs two integer scores in $[0,5]$ corresponding to global structural alignment and fine-grained structural and textural alignment.

We average global and fine-grained scores to obtain video-level scores, reflecting the overall alignment of the generated video with the target attraction.

\paragraph{Human Evaluation.}
We additionally conduct human evaluation following the same protocol as VLM-as-A-Judge. Human annotators are provided with the ground-truth reference image and the same uniformly sampled frames from the generated video. Annotators are asked to assess alignment along the same two dimensions as VLM, using an integer scale from 0 to 5. This human evaluation serves as a complementary benchmark for validating the reliability of the proposed automatic metrics and the VLM-based judgments.

\section{Experiment Setup}
We conduct experiments on \dataset. For each attraction, we generate a single video using the OpenAI Sora 2 \citep{openai_sora2_2025} model conditioned on the corresponding detailed instructions. All videos are generated in portrait orientation with a fixed duration of 4 seconds to ensure consistent presentation across samples.

For evaluation, Patch-level CLIP is computed using the DINOv2-Large\footnote{\texttt{facebook/dinov2-large}} \citep{oquab2023dinov2}. For keypoint-based structural alignment, landmark-relevant regions are detected using GroundingDINO\footnote{\texttt{IDEA-Research/grounding-dino-base}} \citep{liu2023grounding}  and segmented with SAM\footnote{\texttt{facebook/sam-vit-huge}} \citep{kirillov2023segment}. Dense keypoint correspondences are extracted using LoFTR \citep{sun2021loftrdetectorfreelocalfeature} in outdoor mode, which is well suited for large-scale scene and landmark matching. We set $\tau=3000$ and $\beta=1.5$ based on empirical validation.

\section{Findings and Analysis}

\subsection{Metric Validation}

\begin{wrapfigure}{r}{0.5\columnwidth}
\centering
\includegraphics[width=0.5\columnwidth]{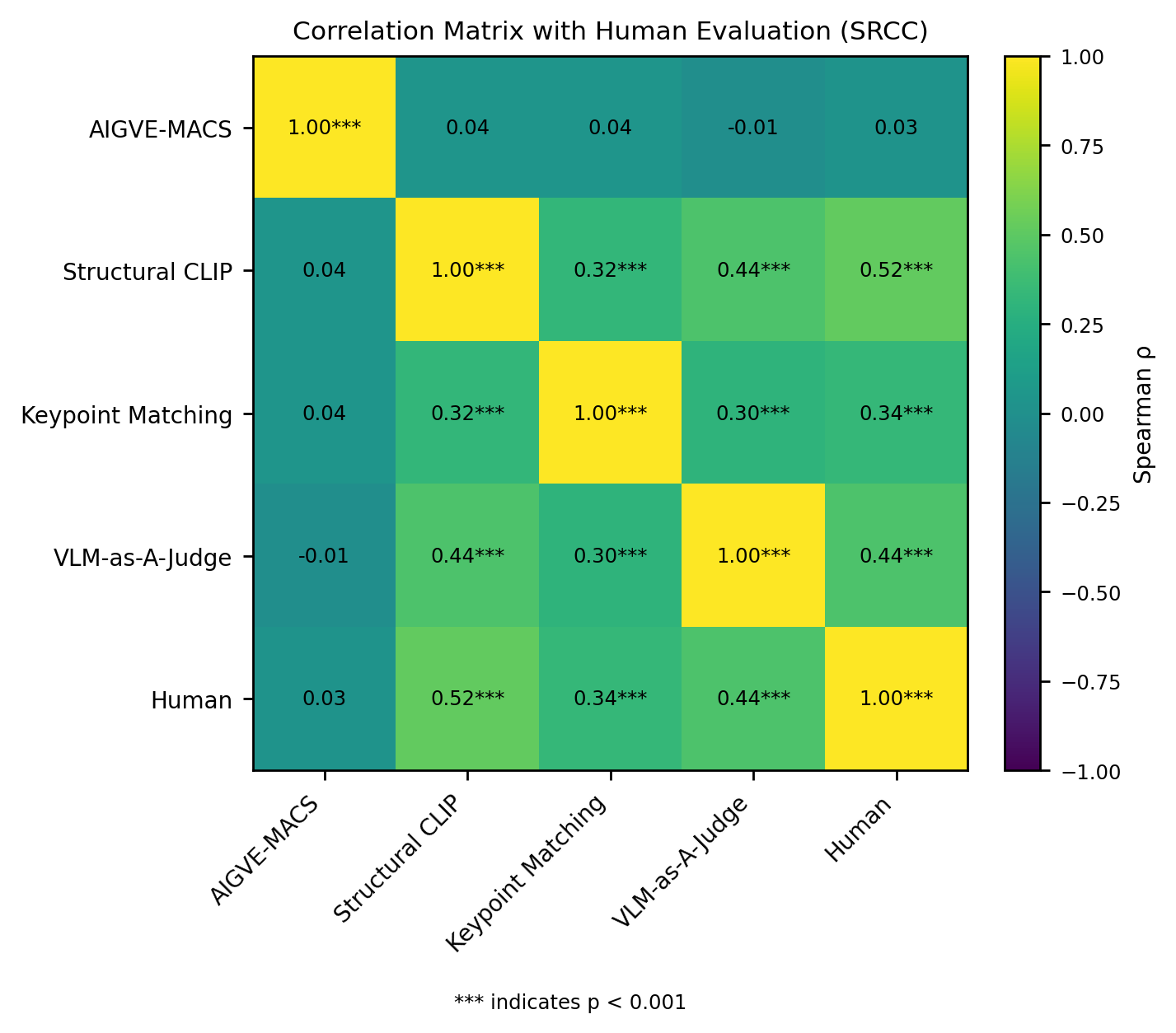}
\vspace{-25pt}
\caption{Spearman rank correlation coefficient between human evaluation scores and automatic metrics.}
\label{fig:srcc}
\vspace{-25pt}
\end{wrapfigure}

We first examine the alignment between our automatic metrics and human judgments. Figure~\ref{fig:srcc} reports the Spearman rank correlation coefficient (SRCC) between human evaluation scores and the metrics used in our framework.

As shown in the figure, AIGVE-MACS exhibits little to no correlation with the other metrics or with human knowledge scores, indicating that it primarily captures video generation quality rather than attraction-specific visual knowledge. This result confirms that our evaluation successfully disentangles quality-related factors from knowledge-oriented assessment.

In contrast, all knowledge-oriented metrics, Patch-level CLIP, Keypoint-Based Local Alignment, and VLM-as-A-Judge, exhibit significant positive correlations with human evaluations. At the same time, the correlations among these automatic metrics themselves are moderate, indicating that they are not redundant. This pattern suggests that each metric captures a distinct yet complementary aspect of attraction-specific visual knowledge. Together, their consistent but differentiated alignment with human judgments demonstrates that the proposed metrics jointly provide a comprehensive and complementary evaluation of the global visual knowledge of video generation models.

\subsection{Geo-Knowledge Disparity Across Popularity}


\begin{figure}[t!]
    \centering
    \includegraphics[width=0.90\linewidth]{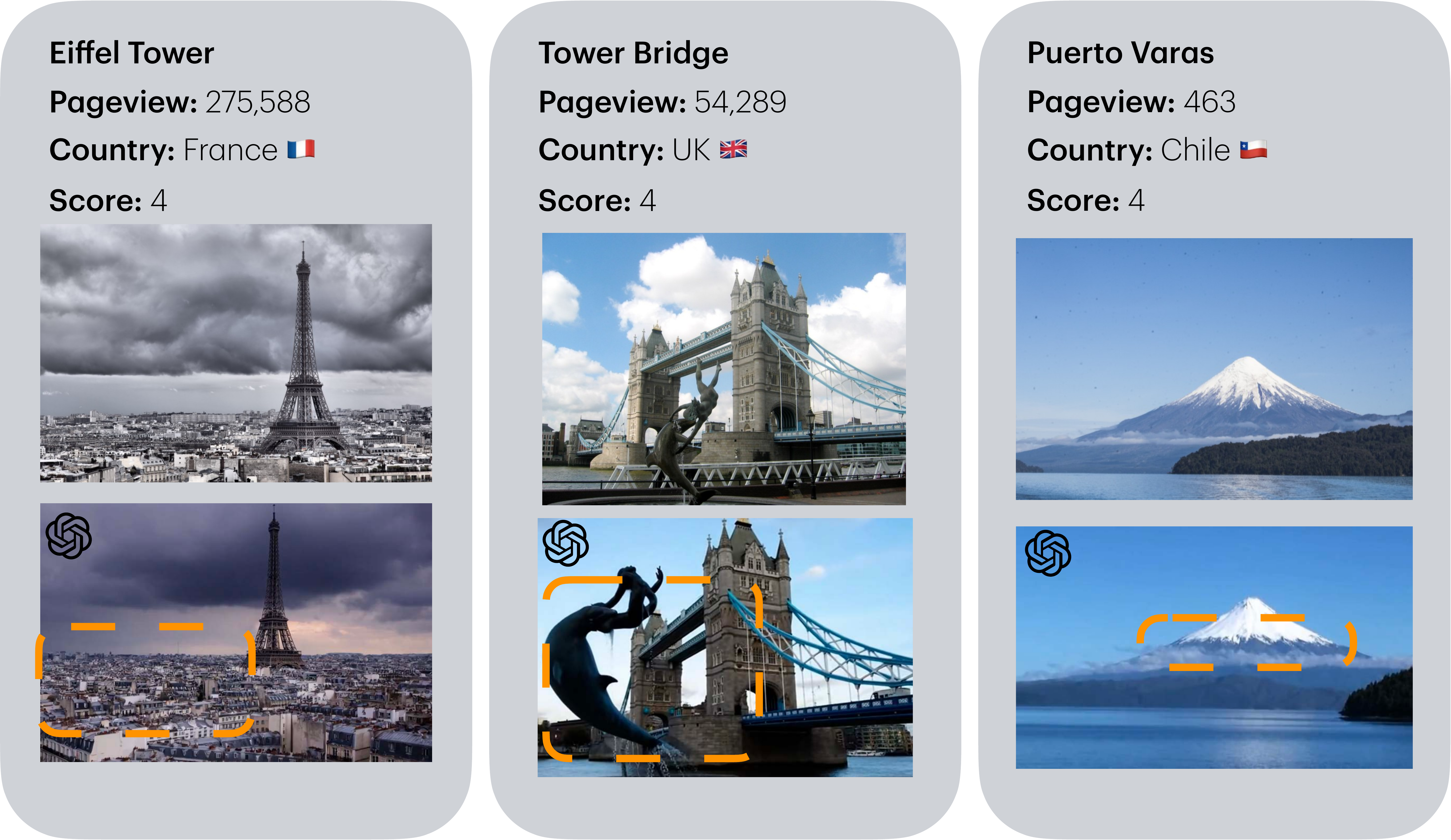}
    \caption{Qualitative case studies illustrating model performance across attractions with varying popularity. For each attraction, the top image shows the ground-truth reference, while the bottom image is generated by the video generation model. Orange dashed outlines highlight regions where the generated content deviates from the reference, indicating structural or geometric errors.}
    \label{fig:case_study}
\end{figure}

\begin{table}[t]
\centering
\small
\scalebox{0.95}{
\begin{tabular}{l|ll|l}
\toprule
\textbf{Metric} 
& \textbf{SRCC} 
& \textbf{$p$-value} 
& \textbf{Linear Slope} \\
\midrule
AIGVE-MACS 
& $-1.94 \times 10^{-2}$ 
& $6.82 \times 10^{-1}$ 
& $-2.35 \times 10^{-7}$ \\

Patch-level CLIP 
& $2.47 \times 10^{-1}$ 
& $1.20 \times 10^{-7}\,^{***}$ 
& $4.73 \times 10^{-7}\,^{***}$ \\

Keypoint Matching 
& $9.63 \times 10^{-2}$ 
& $4.14 \times 10^{-2}\,^{*}$ 
& $1.62 \times 10^{-7}$ \\

VLM-as-A-Judge 
& $1.93 \times 10^{-1}$ 
& $3.86 \times 10^{-5}\,^{***}$ 
& $4.42 \times 10^{-6}\,^{***}$ \\

Human Evaluation 
& $2.30 \times 10^{-1}$ 
& $8.90 \times 10^{-7}\,^{***}$ 
& $5.39 \times 10^{-6}\,^{***}$ \\
\bottomrule
\end{tabular}
}

\caption{Trend analysis between attraction popularity and alignment scores. We report SRCC, two-sided $p$-values, and linear regression slopes. Significance stars indicate $p$-values for the corresponding statistic: $^{*}p<0.05$, $^{**}p<0.01$, $^{***}p<0.001$.}
\label{tab:popularity_trend}
\end{table}

Our first analysis examines whether a model’s geographically grounded visual knowledge is influenced by attraction popularity, which serves as a proxy for the amount of visual data available during training. Intuitively, more popular attractions are more frequently photographed and documented online, and thus are likely to be better represented in the training corpus. We measure attraction popularity using Wikipedia page view statistics. To quantify the relationship between popularity and visual knowledge alignment, we compute the SRCC, which captures monotonic trends, as well as the slope of a linear regression, which reflects the magnitude of alignment score changes as popularity increases.

As shown in Table~\ref{tab:popularity_trend}, all knowledge-oriented metrics exhibit weak correlations with attraction popularity. Besides, their magnitudes are small, and the corresponding regression slopes are close to zero, indicating that increased popularity leads to only marginal gains in alignment. The qualitative examples in Figure~\ref{fig:case_study} further support this finding: attractions with vastly different popularity levels exhibit comparable levels of visual fidelity, with errors typically localized rather than a global disparity. Together, these results suggest that the model maintains \textbf{a relatively uniform level of visual knowledge across attractions with varying popularity}, and that the visual knowledge expressed by Sora2 is not strongly driven by differences in training data volume associated with attraction popularity.







\begin{wraptable}{r}{0.40\columnwidth}
\vspace{-6pt}
\centering
\setlength{\tabcolsep}{6pt}
\renewcommand{\arraystretch}{1.15}
\scalebox{0.90}{
\begin{tabular}{l|ccc}
\toprule
\textbf{Region} & \textbf{$N$} & \textbf{Mean} & \textbf{Std} \\
\midrule
Africa   & 8   & 3.00 & 0.71 \\
Americas & 67  & 3.62 & 0.77 \\
Asia     & 52  & 2.93 & 0.91 \\
Europe   & 249 & 3.02 & 0.86 \\
Oceania  & 5   & 3.60 & 0.42 \\
\midrule
Global East  & 147 & 3.03 & 0.83 \\
Global West  & 302 & 3.13 & 0.87 \\
\midrule
Global South & 178 & 3.00 & 0.81 \\
Global North & 271 & 3.16 & 0.89 \\
\bottomrule
\end{tabular}
}

\caption{Human alignment scores by region.}
\label{tab:human_identity_continent}
\vspace{-15pt}
\end{wraptable}

\subsection{Geo-Knowledge Disparity Across Regions}
\label{subsec:continent_analysis}

Following \cite{lalai2025worldaccordingllmsgeographic}'s work on geographic bias analysis, we examine whether the model’s geographically grounded visual knowledge varies meaningfully across regions. We group attractions by continent and further analyze disparities along two complementary axes: Global North vs. Global South (GN/GS), reflecting differences in economic development and data availability, and Global West vs. Global East (GW/GE), capturing broad cultural distinctions.

Table~\ref{tab:human_identity_continent} reports the mean and standard deviation of human alignment scores by continent. Across regions, average scores are broadly similar, with comparable variability within each group. Higher variance observed in Africa and Oceania is attributable to their smaller sample sizes.

To assess practical significance, we conduct pairwise bootstrap equivalence testing with a tolerance of $\delta = 1.0$. We choose this threshold because the alignment scores are measured on an integer scale from 0 to 5, and a one-point difference corresponds to a perceptually meaningful change in alignment quality. As shown in Table~\ref{tab:equivalence_continent}, all continent-pair differences fall within the equivalence range, indicating largely consistent alignment scores across continents.

We further observe that comparisons along the GN/GS and GW/GE axes yield small mean differences, with confidence intervals fully contained within the equivalence margin. These results suggest that differences along both the development axis (GN vs. GS) and the cultural axis (GW vs. GE) are practically small.

Overall, these findings indicate that the model’s \textbf{geographically grounded visual knowledge is relatively uniform across continents}, development levels, and cultural groupings, with no evidence of large, systematic regional disparities under a one-point tolerance threshold.

\subsection{Impact of Instruction Specificity}

\begin{table}[]
   
\centering
\small
\scalebox{1}{
\begin{tabular}{l|ccc}
\toprule
\textbf{Region Pair} & \textbf{Mean Diff.} & \textbf{95\% CI} & \textbf{Eq.} \\
\midrule
Americas vs Asia   & 0.69 & [$0.38$, $0.99$] & \checkmark \\
Americas vs Europe & 0.60 & [$0.38$, $0.81$] & \checkmark \\
Europe vs Asia     & 0.09 & [$-0.17$, $0.36$] & \checkmark \\
\midrule
Global West vs East  & 0.09 & [$-0.07$, $0.26$] & \checkmark \\
Global North vs South & 0.17 & [$0.01$, $0.32$] & \checkmark \\
\bottomrule
\end{tabular}
}

\caption{Pairwise equivalence testing of regional differences in human alignment scores using bootstrap confidence intervals with a tolerance of $\delta = 1.0$. A checkmark indicates practical equivalence.}
\label{tab:equivalence_continent}
\end{table}

We examine how prompt specificity affects the visual knowledge expressed by video generation models using a subset of 50 attractions. Each attraction is paired with two prompts: a short, single-sentence instruction and a detailed, multi-sentence instruction describing the same ground-truth image. All other generation and evaluation settings are held constant.

Table~\ref{tab:caption_detail_comparison} presents a paired comparison between videos generated from short and detailed prompts. Across all knowledge-oriented metrics, detailed prompts consistently yield higher alignment scores, suggesting better global and fine-grained alignment from a semantic perspective. However, the corresponding effect sizes are small to moderate, indicating that while increased prompt specificity consistently improves alignment, it does not fundamentally alter the model’s underlying visual knowledge. Overall, these results suggest that richer textual instructions facilitate more accurate expression of visual knowledge, while reinforcing our earlier finding that \textbf{geographically grounded visual knowledge remains relatively stable and not overly sensitive to prompt verbosity.}
\begin{table}[t]
\centering
\small
\setlength{\tabcolsep}{6pt}
\scalebox{1}{
\begin{tabular}{l|ccc|c}
\toprule
\textbf{Metric} 
& $\textbf{Mean}_S$ 
& $\textbf{Mean}_D$ 
& \textbf{$p$} 
& \textbf{$d$} \\
\midrule
S-CLIP 
& 0.66
& 0.68 
& 0.00027***
& 0.57 \\
Kpt-Almt
& 0.44 
& 0.47 
& 0.041*
& 0.26 \\
VLM
& 3.49 
& 3.78 
& 0.017*
& 0.33 \\
\bottomrule
\end{tabular}

}

\vspace{0.3em}
\caption{
Paired comparison between videos generated from short and detailed captions.
Mean$_S$ and Mean$_D$ denote average scores under short and detailed caption settings, respectively.
Statistical significance is assessed using the Wilcoxon signed-rank test ($p$), and effect size is measured by Cohen's $d$.
S-CLIP denotes Structural CLIP, Kpt-Almt denotes keypoint-based local alignment, and VLM denotes the VLM-as-A-Judge score. $^{***}p<0.001$, $^{*}p<0.05$.
}

\label{tab:caption_detail_comparison}
\end{table}

\subsection{Discussion and Future Directions}

A counterintuitive finding of this work is that, unlike LLMs \citep{Manvi2024Large, Decoupes2024Evaluation}, text-to-video models exhibit relatively stable geographically grounded visual knowledge across regions. This stability persists across attraction popularity levels, suggesting that visual knowledge is not strongly affected by variations in training data associated with specific locations, in contrast to patterns commonly observed in LLMs.

One possible explanation is that current text-to-video models have not yet reached expert-level visual knowledge. Across our experiments, most alignment scores fall in a medium range (approximately 3--4 on a 5-point scale), indicating relatively uniform underfitting across regions. As these models continue to improve, geographic disparities may become more pronounced, similar to trends observed in LLMs.

Another hypothesis relates to differences in generation mechanisms~\citep{prabhudesai2025diffusionbeatsautoregressivedataconstrained}. Autoregressive LLMs generate content sequentially, where early errors can accumulate and amplify biases. In contrast, diffusion-based video models iteratively refine samples through denoising, allowing continuous correction during generation, which may naturally mitigate error accumulation and reduce disparity.

These findings motivate future work comparing knowledge disparities between diffusion-based and autoregressive language models, as well as longitudinal analyses tracking how geographic uniformity evolves as text-to-video models scale. Overall, the relatively uniform visual knowledge observed in current text-to-video models suggests they are promising candidates for applications requiring broad, global visual coverage.

\section{Conclusion}

In this work, we studied the geo-equity and geographically grounded visual knowledge of modern text-to-video models through their generated outputs. We introduced Geo-Attraction Landmark Probing(\ours), a structured evaluation framework that uses tourist attractions as geographically grounded proxies for regional visual knowledge and evaluates generated videos with complementary quality- and knowledge-oriented metrics. By grounding evaluation in real-world reference imagery and applying a consistent evaluation protocol across diverse regions, GAP enables systematic analysis of whether a model’s visual knowledge is expressed uniformly across the globe.

\bibliography{colm2026_conference, custom, aigve_score}
\bibliographystyle{colm2026_conference}

\appendix
\newpage
\section{Prompts}

\subsection{Instruction Generation Prompt}
\label{sup:caption_prompt}

\begin{tcolorbox}[
    breakable,
    colback=gray!2,
    colframe=gray!50,
    boxrule=0.7pt,
    enhanced,
    sharp corners,
    title=\textbf{Instruction Generation Prompt}
]

\begin{ttfamily}\small
You are a vision-language model tasked with analyzing an image and producing structured
captions conditioned on a known landmark.

\medskip

\textbf{Task:}
Analyze the provided image and produce a JSON object with three fields:
\begin{enumerate}
    \item \texttt{"is\_correct\_place"}: A Boolean indicating whether the image matches the expected landmark.
    \item \texttt{"detailed\_caption"}: A cinematic caption of 3--6 sentences describing the scene.
    \item \texttt{"short\_caption"}: A single-sentence summary of the scene.
\end{enumerate}

\medskip

\textbf{Context:}
The expected landmark is \texttt{"<landmark\_name>"} in \texttt{<city>}, \texttt{<country>}.

\medskip

\textbf{Instructions:}
\begin{enumerate}
    \item First, determine whether the image depicts \texttt{<landmark\_name>} in \texttt{<city>}, \texttt{<country>}.
    \begin{itemize}
        \item If \textbf{YES}, set \texttt{"is\_correct\_place"} to \texttt{true} and generate both captions.
        \item If \textbf{NO}, set \texttt{"is\_correct\_place"} to \texttt{false} and output:
        \begin{itemize}
            \item \texttt{"detailed\_caption"}: \\
            \texttt{"This image is not <landmark\_name> in <city>, <country>. It appears to be <identify landmark if possible>."}
            \item \texttt{"short\_caption"}: \\
            \texttt{"This is not <landmark\_name> in <city>, <country>."}
        \end{itemize}
        Do not generate any additional descriptions.
    \end{itemize}

    \item If the image \emph{is} \texttt{<landmark\_name>} in \texttt{<city>}, \texttt{<country>}, generate:
    \begin{itemize}
        \item \textbf{Detailed caption (3--6 sentences)} including:
        \begin{itemize}
            \item Aspect ratio (e.g., 16:9, 9:16)
            \item Composition (camera angle, framing, foreground/background, depth)
            \item Location of the main landmark(s) within the frame
            \item Weather, lighting, and overall tone
            \item Environmental and architectural details
            \item Cinematic or stylistic cues useful for video generation
        \end{itemize}
        \item \textbf{Short caption (1 sentence)} summarizing the same scene.
    \end{itemize}

    \item Writing constraints:
    \begin{itemize}
        \item Do not mention verification, checking, or internal reasoning.
        \item Do not refer to ``this image'' or ``the viewer''.
        \item Captions must be visual, concrete, and suitable for scene recreation.
        \item Output must be valid JSON.
    \end{itemize}
\end{enumerate}

\medskip

\textbf{Output Format:}
Return \textbf{only} a JSON object with the following structure:
\begin{verbatim}
{
  "is_correct_place": true/false,
  "detailed_caption": "<detailed caption>",
  "short_caption": "<1 sentence caption based on detailed caption>"
}
\end{verbatim}

\end{ttfamily}

\end{tcolorbox}

\end{document}